# Controlling Search in Very Large Commonsense Knowledge Bases: A Machine Learning Approach

Abhishek Sharma[1]  Michael J. Witbrock[2]  Keith M. Goolsbey[1]

[1]Cycorp, Inc., 7718 Wood Hollow Drive, Suite 250, Austin, TX 78731
[2]Lucid AI, 600 Congress Avenue, Austin, TX 78701
abhishek@cyc.com  witbrock@lucid.ai  goolsbey@cyc.com

## Abstract

Very large commonsense knowledge bases (KBs) often have thousands to millions of axioms, of which relatively few are relevant for answering any given query. A large number of irrelevant axioms can easily overwhelm resolution-based theorem provers. Therefore, methods that help the reasoner identify useful inference paths form an essential part of large-scale reasoning systems. In this paper, we describe two ordering heuristics for optimization of reasoning in such systems. First, we discuss how decision trees can be used to select inference steps that are more likely to succeed. Second, we identify a small set of problem instance features that suffice to guide searches away from intractable regions of the search space. We show the efficacy of these techniques via experiments on thousands of queries from the Cyc KB. Results show that these methods lead to an order of magnitude reduction in inference time.

## 1 Introduction

Commonsense reasoning has always been a core problem for artificial intelligence (AI) systems. Effective reasoning about the external world often involves drawing deductively valid conclusions from known facts. Unfortunately, given the combinatorial explosiveness of reasoning in expressive knowledge-based systems (KBS), even simple queries might get "lost" in millions of seemingly relevant inference paths. Efficient reasoning in such systems is critical for building large-scale AI systems.

Ordering heuristics play an important role in optimization of reasoning in KBS for at least two reasons: First, inference algorithms of KBS (e.g., backward chaining [Russell and Norvig 2003] in Cyc, tableaux algorithms in description logic (DL)) typically represent the search space as a graph, the structure of which is determined by the rules applicable to the given node in the graph. Generally, many rules might simultaneously apply to a given vertex, and the order of rule expansion can have a significant effect on efficiency [Tsarkov and Horrocks 2005]. Second, researchers have used first-order logic (FOL) theorem provers as tools for inference with very expressive languages (e.g., OWL DL, the Semantic Web Rule Language (SWRL)) where reasoning with the complete language is beyond the scope of existing DL algorithms or the language does not correspond to any decidable fragment of FOL [Tsarkov *et al.* 2004, Horrocks and Voronkov 2006]. Very large FOL systems often have thousands to millions of axioms, of which only a few are relevant for answering any given query [Hoder and Voronkov 2011, Tsarkov *et al.* 2004]. Since hundreds of thousands of axioms that are irrelevant for a given query might overwhelm resolution-based theorem provers, the reasoner is expected to assess the utility of further expanding numerous incomplete inference paths. A naïve ordering of paths can lead to potentially infinite subtrees and cause unproductive backtracking.

To make inferences more efficient, this paper suggests two types of ordering heuristics. First, we discuss how implausible search paths are created when domain-specific axioms are used to prove queries involving fairly general predicates. We argue that decision trees can be used to represent the semantic context in which a rule is likely to contribute to a proof. We show that ordering nodes with the help of decision trees helps in guiding search toward the solution. Second, a key impediment in the development of fast broad-application first-order reasoning systems has been an insufficient understanding of what makes problems difficult. We propose a comprehensive set of features that correlate with the answerability of nodes. We run the inference engine on a large number of queries, sample nodes from the resulting search graphs, and record values for their instance features. We use statistical regression methods to derive a model for predicting the answerability of nodes. We use the resulting model to order nodes during search and demonstrate that this improves search performance.

This paper is organized as follows: We start by discussing relevant previous work. Our decision tree algorithm and statistical regression methods are discussed next. We conclude by discussing our results and plans for future work.

## 2 Related Work

Prior research has examined the use of machine learning to identify best heuristics[1] for problems [Bridge *et al*. 2014] and to select a small set of axioms/lemmas that are most relevant for answering a set of queries [Hoder and Voronkov 2011, Sharma and Forbus 2013, Meng and Paulson 2009, Kaliszyk *et al*. 2015, Kaliszyk and Urban 2015, Alama *et al.* 2014]. In contrast, we focus on ordering heuristics that enable inference algorithms to reason with *all* axioms. In [Taylor *et al*. 2007], the authors use reinforcement learning to guide inference, whereas in [Tsarkov and Horrocks 2005], the authors study different types of rule-ordering heuristics (e.g., preference between ∃ and ⊔ rules) and expansion-ordering heuristics (e.g., descending order of frequency of usage of each of the concepts in the disjunction). This paper proposes that rule-ordering heuristics should be based on the *search state*, and we use a regression-based model to learn the effects of different features on the answerability of nodes. Work in other fields (e.g., database community [Chaudhuri 1998], SAT reasoning [Hutter *et al.* 2014], answer set programming [Brewka *et al*. 2011]) is less relevant because the studies do not address the complexity of deep and cyclic search graphs that arise from expressive first-order reasoning. To the best of our knowledge, no work in the AI community has used decision trees and statistical regression-based methods to control inference in large commonsense reasoning systems.

## 3 Background

We assume familiarity with the Cyc representation language [Lenat and Guha 1990, Matuszek *et al.* 2006, Taylor *et al.* 2007]. In Cyc, concept hierarchies are represented by the 'genls' relation. For instance, (genls Person Mammal) holds. During backward inference, the rule $P(x) \rightarrow Q(x)$ is used to *transform* a query like Q(a) into P(a). The link between Q(a) and P(a) is a type of *transformation link.* A node like $s_0$: (and (performedBy ?x JohnMcCarthy-ComputerScientist) (isa ?x Buying)) leads to sub-goals like $s_1$: (isa JohnMcCarthyBuysABook-012 Buying) and $s_2$: (isa JohnMcCarthyWritesAPaper-087 Buying), some of which may be satisfiable. The links between $s_0$ and $s_1$ and $s_0$ and $s_2$ are examples of *restriction links*. Transformation and restriction links play a major role in determining the out-degree of nodes. Every node in the search graph is timestamped with an *id*. A node *y* is called a *successor* of *x* if there is a path consisting of transformation links from *x* to *y* and $id(x) < id(y)$. A node *x* is a *parent* of *y* if a transformation link exists between *x* and *y* and $id(x) < id(y)$. *Parents*(*x*) and *Successors*(*x*) denote the sets of all parents and successors of node *x* respectively. Let S be the set of all nodes in a search graph. Then, a *transformation link set* p = {a(1), a(2), …,a(n)} is a set of transformation links that transform an initial state $s_0$ to an intermediate state $s_n$. *Rule(a)* and *Substitutions(a)* denote the rule and bindings associated with the transformation link *a*. Transitive inference is well supported in Cyc. The query (genls ?x Person) has more than 6,700 answers because the predicate 'genls' allows transitive inference in its first argument position. The aforementioned query has one *open transitive argument position*.

Reasoning in Cyc KB is difficult due to the sheer size of the KB and the expressiveness of the CycL representation language. In its default inference mode, the Cyc inference engine uses the following types of axioms/facts during backward inference: (i) 21,743 role inclusion axioms (e.g., $P(x, y) \rightarrow Q(x, y)$), (ii) 2,601 inverse role axioms (e.g., $P(x, y) \rightarrow Q(y, x)$), (iii) 365,593 concepts and 986,965 concept inclusion axioms (i.e., 'genls' facts), (iv) 817 transitive roles, (v) 99,238 complex role inclusion axioms (e.g., $P(x, y) \wedge Q(y, z) \rightarrow R(x, z)$), and (vi) 31,897 binary roles and 7,980 roles with arities greater than two. The KB has 21.7 million assertions and 652,037 individuals. To control search in such a large KBS, inference algorithms often use different control strategies. They distinguish between a set of clauses known as the *set of support*[2] that define the important facts about the problem and a set of *usable axioms* that are outside the set of support (e.g., see the OTTER theorem prover [Russell and Norvig 2003]). At every step, such theorem provers resolve an element of the set of support against one of the usable axioms. To perform best-first search, a heuristic control strategy mesures the "weight" of each clause in the set of support, picks the "best" clause, and adds to the set of support the immediate consequences of resolving it with the elements of the usable list [Russell and Norvig 2003]. Cyc uses a set of heuristic modules to identify the best clause from the set of support. A *heuristic module* is a tuple $h_i = (w_i, f_i)$, where $f_i$ is a function $f_i: S \rightarrow \mathbb{R}$ that assesses the quality of a node, and $w_i$ is the weight of $h_i$. The net score of a node *s* is $\Sigma_i w_i f_i(s)$, and the node with the highest score is selected for further expansion. In next two sections, we discuss two heuristic modules for focusing search.

## 4 Decision Trees for Focused Search

The basic idea behind this approach is best explained with a few examples. Consider the rules shown below[3]:

(sitTypeIsSpecWithTypeRestrictionOnRolePlayer ?absorption PhotonAbsorption absorber ?type) ∧ (sitTypeIsSpecWithTypeRestrictionOnRolePlayer ?excitation ChemicalObjectExcitation objectOfStateChange ?type)
→ (cotemporalProperSubEventTypes ?absorption ?excitation)    **(Rule A1)**

(objectFoundInLocation ?ARG1 ?ARG2) ∧ (geopoliticalSubdivision ?OTHER ?ARG2) → (objectFoundInLocation ?ARG1 ?OTHER)    **(Rule A2)**

To answer the query (properSubEventTypes BirthdayParty ?x), an inference engine would backchain on rule A1 (see footnote

---

[1] Examples of heuristics (or strategies) include "give priority to axioms in clause selection" and "sort symbols by inverse frequency".

[2] For instance, the negated query is often used as the set of support.

[3] A sentence of the form (sitTypeIsSpecWithTypeRestrictionOnRolePlayer SPEC SIT-TYPE ROLE TYPE) means that SPEC is the unique specialization of SIT-TYPE, a specialization of 'Situation', such that all objects that play ROLE in instances of SPEC are instances of TYPE.

3) because 'cotemporalProperSubEventTypes' is a sub-role of 'properSubEventTypes'. This transformation would lead us to reason about photon absorption[4]. Similarly, we would backchain on rule A2 to answer the query (objectFoundInLocation ?x MesophyllCell-001). Such search paths are unlikely to succeed.

General knowledge bases often have heavily used predicates with hundreds of specializations. These specializations partition the space into several domains. For example, while rule A1 is expected to be useful for naïve physics, A2 is expected to be useful when reasoning about geographic sites. Implausible search paths arise when a mismatch exists between the query and the implied context in which an axiom is likely to work. In this paper, we suggest that type/concept based decision trees are the right representation choice for this problem because rules are expected to fire for a certain class (or type) of things. Therefore, we associate *restrictive information* with the variables of axioms. Although the variables are expected to range over their entire domain, the restrictive information specifies a subset of the domain over which the rule has been observed to work. These restrictions are specified in terms of *sorts* or concepts. They derive from the results of successful uses of the given rule. A small set of successful bindings for rule 2 is shown in Table 1. The fact that Minneapolis, Anaheim, and Rochester are US cities helps us derive the *sorted generalization* [Page and Frisch 1992] that ?ARG2 is likely to range over the set USCity. Formally, a restriction condition is a pair, $x{:}\tau$, where $x$ is a variable and $\tau$ is a concept. Let $\Sigma$ denote the set of sentences that represent relationships among the concepts. Then, a substitution $\theta$ satisfies the restriction condition $x{:}\tau$ if it maps $x$ to a ground term $t$ and $\Sigma \models \tau(t)$. The $j^{th}$ restriction condition for axiom $a$, $RC(a, j)$, can be represented as $\bigwedge_{x(i)\in Vars(a)} x(i){:}\tau(i)$, where $Vars(a)$ is the set of variables in $a$. A disjunction of such constraints can be specified as $\bigvee_i RC(a,i)$, and decision trees are a natural representation for such constraints. Our algorithm for constructing a decision tree from a set of successful rule bindings is shown in Figure 1. The compact decision tree (induced from 1900 bindings) for rule 2 is simply:

**?ARG2:**GeopoliticalEntity $\wedge$ **?OTHER:**GeographicalRegion $\wedge$ **?ARG1:**TerrestrialFunctioningObject

| ?OTHER | ?ARG2 | ?ARG1 |
|---|---|---|
| Minnesota-State | CityOfMinneapolisMN | UnivOfMinnesota |
| NewYork-State | CityOfRochesterNY | Ginna-NuclearPowerPlant |
| California-State | CityOfAnaheimCA | AngelStadiumOfAnaheim |

**Table 1: Partial Training Set for Rule 2**

The algorithm *CreateTree* (shown in Figure 1) takes as input a training set and the variables that occur in the rule. The training set is generated by querying the antecedent of the rule for a fixed duration of time. The bindings returned by the query results form the *TrainingSet*. Given a tuple from the training set (see Table 1), we compute the generalizations of the bindings.[5] In step 3 of the algorithm, membership in the most specific maximally covering generalization is chosen as the branching test. When a tuple satisfies this test, we explore constraints for other variables in the AND branch (step 6). Otherwise, other values for the variable are considered in the OR branch (step 7). We stop growing the tree (step 2) when the number of unexplained training examples is less than a pre-determined fraction of the full training set. The complexity of top-down decision tree induction is $O(m^2 \cdot n)$ where $m$ is the number of attributes, and $n$ is the size of the training set [Kent and Hirschberg 1996]. To use decision trees during search we can define a heuristic module with the following function, $f_{DT}(s)$, for assessing the quality of nodes:

$$\sum_{p\in L(s)} \sum_{a(i)\in p} \frac{1}{|p|} I(Substitutions(a(i)), Tree(Rule(a(i)))) \quad ..(1)$$

In (1), L(s) is the set of all transformation link sets from node $s$ to the root, and $I(\theta, tree)$ is 1 when the substitution $\theta$ satisfies the restriction conditions specified by the tree, and 0 otherwise[6]. This module will prioritize search paths that satisfy the restriction conditions. For instance, the path that uses rule A2 to answer the query (objectFoundInLocation ?x MesophyllCell-001) would be expanded late because MesophyllCell-001 is not transitively a GeographicalRegion (see the constraint for variable ?OTHER). This helps in early evaluation of inference steps that use rules from the domains that are pertinent for the given query. However, since the KB might have thousands of rules that are relevant for a query, we also need other ways to steer the search toward more productive states. In the next section, we propose a statistical approach to solving this problem.

---

**Input**: *TrainingSet*, a set of tuples of successful bindings for the rule.
   *ListOfVars*, the list of variables used in the rule.
**Output**: Decision Tree for the rule.

1. Create a *Root* node for tree.
2. If stopping criterion has been reached then return *Root*.
3. (*var, value*) ← the variable and value that provide the best covering generalization
4. *AccountedSet* ← Subset of *TrainingSet* that have *var = value*
5. *UnaccountedSet* ← *TrainingSet – AccountedSet*
6. *LeftChild* ← CreateTree (*AccountedSet, ListOfVars – var*)
7. *RightChild* ← CreateTree(*UnaccountedSet, ListOfVars*)
8. Add *LeftChild* as a new branch below *Root* corresponding to the test *var = value*.
9. Add *RightChild* as a new branch below *Root* corresponding to the test *var ≠ value*.
10. Return *Root*.

**Figure 1: The CreateTree Algorithm**

---

[4] Readers might wonder about domain constraints. The first argument to 'sitTypeIsSpecWithTypeRestrictionOnRolePlayer' is expected to be a specialization of 'Situation', and the concept 'BirthdayParty' satisfies this condition. The generality of some domain constraints ensures that it is difficult to identify implausible sub-queries.

[5] The generalization of a substitution, s, is the set Gen(s) = {c | $\Sigma \models$ (isa s c)}. For instance, PopulatedPlace is a generalization of Minnesota-State.

[6] The 'Tree' function in (1) returns the decision tree for a given rule.

# 5 Statistical Meta-Search Learning

Is it possible to predict whether an inference engine will be able to solve an arbitrary node generated during search? In this section, we show how supervised machine learning methods can be used to build models that predict the number of answers for a problem instance. Such models can be used by the inference engine to decide how to allocate computational resources. Moreover, by shedding light on the sources of hardness in problem instances, they help in improving knowledge representation and fuel development of new algorithms.

---

(1-8) *Problem size and type features (Cheap)*: Number of variables, literals, fully unbound literals, single/**multi literal** query, fully/partially bound query, **fully unbound single literal query**.

(9-13) *Problem state features (Cheap)*: Depth, number of transformation and restriction links, **potential fan-out score**, number of rules used recursively in reaching the state.

(14-18) *Knowledge level features (Moderate)*: Generality estimate of unbound literals, min $_{TERMS}$ TermGenerality(t), **number of GAFs for predicate in single literal query**, min $_P$ NumGafs(p), Generality estimate of predicate in fully unbound single literal query.

(19-21) *Transitivity features (Moderate)*: Number of open transitive argument positions, Number of open transitive argument positions in queries with multiple variables, Number of open argument positions in 'genls' and 'disjointWith' literals.

(22-29) *Probing features (Expensive)*: Number of transformation links (mean), number of literals (mean), out degree of nodes (median and max), **number of variables (median)**, $|\cup_{p \in Parents\,(node)}\{s\,|s \in Successors(p)\}|$, depth (median), **Knuth's tree size estimate**.

(30-32) *Problem balance features (Cheap)*: Ratio of number of variables and literals, **ratio of number of positive and negative literals, |Number of positive literals -1|**.

(33-38) *Quadratic Terms (Cheap)*: (number of literals)$^2$, (number of free variables)$^2$, depth$^2$, **(number of transformation links)$^2$**, (ratio of number of variables and literals)$^2$, $(|\cup_{p \in Parents\,(node)}\{s\,|s \in Successors(p)\}|)^2$.

(39-44) *Interaction Terms (Moderate)*: depth * (number of transformation links), depth*$|\cup_{p \in Parents\,(node)}\{s\,|s \in Successors(p)\}|$, depth * Knuth's tree size estimate, single literal query * number of open transitive argument positions, min $_{TERMS}$ TermGenerality(t) * single literal query, generality of fully unbound literals *multi literal query.

(45) *Misc. (Cheap)*: Single literal query with procedural support.

(46) *Result feature (Cheap)*: Number of answers

**Figure 2: List of Features and Their Cost of Computation. Features selected by exhaustive subset selection are shown in bold.**

---

To build such models, we take the following steps: (i) *Identification of features*: First, we identify key parameters that represent all known relevant features of problem instances. (ii) *Data collection*: Next, we run the inference engine on a large set of queries and sample nodes from the generated search graph. For each sampled node, the number of answers and a set of *feature values* are recorded. (iii) *Learning*: Finally, we learn a model that maps from instance features to the inference engine's performance, and evaluate it on a test set of queries. After introducing some notation that is used in Figure 2, we discuss each of these steps in detail.

**Notation:** Let P and TERMS denote the set of predicates and terms mentioned in the query respectively. For any predicate *p*, let NumGafs(p) and NumRules(p) denote the number of ground atomic formulas and number of relevant rules for *p*. Moreover, Cyc maintains an estimate of generality of any term based on its position in the ontology. Let TermGenerality (t) denote the generality of any term *t*.

**Feature Identification**: Our features and their cost of computation are shown in Figure 2. Broadly, they can be divided into ten groups. The first group includes well-understood problem size and type features including number of literals and number of variables. The second group contains those attributes that involve examining the path that led to the node. This includes important features that help in maintaining the right shape of the search space. While the feature "depth" is critical in ensuring that the inference engine is not trapped in depth-first infinite regress, the feature "number of transformation links" helps us control the out-degree of nodes. Figures 3 and 4 show the trade-off between depth-first and breadth-first search. We see that the conditional probability of success of a node decreases rapidly with depth. Similarly, Figure 4 shows that most of the successful transformation links are added in the initial phase and the utility of adding an additional transformation link drops rapidly. Table 2 shows the conditional probability of success of nodes as a function of number of literals.

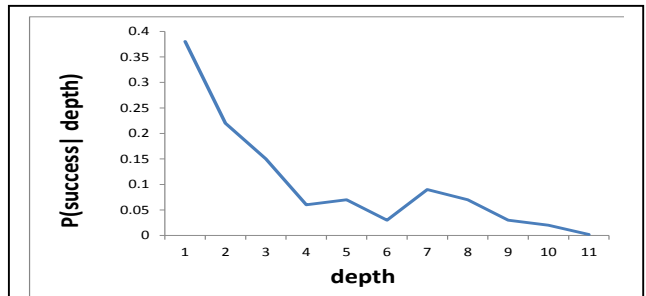

**Figure 3: Likelihood of success as a function of depth.**

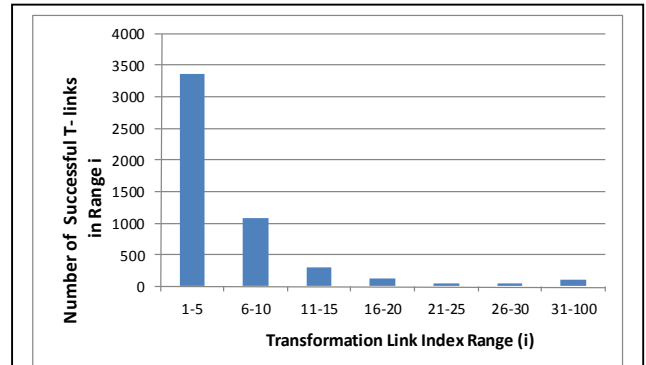

Figure 4: The x-axis shows the index range of transformation links, and the y-axis shows the number of successful transformation links in the given range.

| Number of literals (n) | Prob (success \| n) |
|---|---|
| 1 | 0.940 |
| 2 | 0.009 |
| 3 | 0.001 |
| 4 | 0.003 |
| 5 | 0.003 |
| 6 | 0.003 |

Table 2: Conditional Probability of success as a function of number of literals

The potential fan-out score of a node is a function of the number of rules that can potentially be used with it. Formally it is $\Sigma_P \log_{10} (1+\text{NumRules}(p))$. The next group includes features that encode the level of knowledge the KB has for predicates and terms mentioned in the query. The generality estimate mentioned in the third group is defined as $\prod_P \log_{10} (1+\text{TermGenerality}(p))$. In the fourth group, we include attributes for understanding the cost of transitive queries. The *probing feature* is a special kind of feature that examines the neighborhood of the node to assess its quality. Since locally available information at a node is insufficient for gauging the complexity of the search space below it, in the fifth group, we include features that pick a random path originating at the given node and record descriptive statistics of various properties of interest (e.g., number of literals, out-degree of nodes). For example, the third feature of the fifth group is computed by finding the median out degree of nodes encountered in the randomly selected path. The sixth group captures the balance of the node in two ways: we measure the ratio of number of variables and literals, and the ratio of number of positive and negative literals. Because we expect disjunctive queries to be more difficult, we also note whether non-Horn axioms were used in deriving the state. In the seventh and eighth group, we included quadratic and interaction terms for some salient features.

*Data Collection:* The Cyc KB contains thousands of stored queries of various level of difficulty. We gathered a large amount of data by sampling and running these queries. Forty percent of the nodes from the resulting space were sampled, and the values of the 46 features shown in Figure 2 were recorded. This produced 2.5 million data points.

*Data Transformation and Learning:* Recall that the number of answers is the performance measure, and all other features shown in Figure 2 are predictor features. We performed z-score normalization of the predictor variables by subtracting the mean and dividing the difference by their standard deviation[7]. Given the extreme variability in the number of answers, we use a log-transformation on the result feature (i.e., we predict $\log_{10}(1+$ number of answers$)$). In our initial unpublished work, we experimented with classification techniques (e.g., naïve Bayes, logistic regression) to predict the likelihood of success of a node. Since the results were not very encouraging, we switched to multiple linear regression. In linear regression, the aim is to learn a function of the form $f_{SL}(s) = \Sigma w_i g_i(s)$, where $w_i$ is the weight for the $i^{th}$ feature of node $s$, $g_i(s)$. The function $f_{SL}(s)$ can then be used by a heuristic module to assess the quality of nodes. The values of $w_i$ are determined by minimizing the metric *root mean squared error* (RMSE). The R software was used to estimate the values of $w_i$ [R Core Team 2015]. We used the *repeated random sub-sampling approach* and *10-fold cross validation* to validate our model. While using the former method, 90% of the data was selected at random for training and the rest were used as the test set. This process was repeated 50 times. The mean RMSE from the two validation methods was 0.47 and 0.76 respectively. Multiple $R^2$ and adjusted $R^2$ for this model were equal to 0.76, and the F-statistic was $1.8*10^5$.

For a variety of reasons, the features can be uninformative, correlated or redundant. Therefore, we use *feature selection methods* to identify a small set of features that explains the variance in data as well as the full set of features[8]. Such analysis helps us to identify properties of nodes that strongly affect empirical performance. The set of best 10 features as identified by an exhaustive subset selection method is shown in bold in Figure 2. The $R^2$ value of subset models with these 10 features converged to that of models with all inputs. The presence of features such as "Knuth's tree size estimate" [Knuth 1975], "ratio of number of positive and negative literals" and "(number of transformation links)$^2$" in the selected list suggests the following: (i) Locally available information is insufficient for predicting the complexity of search space, and probing features play an important role in guiding a search. (ii) ensuring that the search graph has the right shape is of critical importance, and reasoners need to find the balance between "depth-first" and "breadth-first" search; (iii) negated literals and disjunctive queries are more difficult to answer. In the next section, we evaluate how these heuristics help the inference engine in answering queries.

## 6 Experimental Results

The selection of benchmark instances for testing the efficacy of heuristics is an important factor in any empirical analysis. Our selection of problem instances was guided by two principles: (i) The benchmark set should consist of queries that are intrinsically difficult to solve for the inference engine. Therefore, we excluded simple queries that can be answered without any backchaining in a few milliseconds (e.g., (isa MarvinMinsky Person), (genls Dog Carnivore)). We focused on queries that needed several transformations (i.e., depth of rule back-chaining) to be answered. (ii) While artificially crafted and randomly generated problem instances are very useful for understanding how syntactic properties affect the behavior of algorithms, the right methodology for generating such instances has not received much attention in the commonsense reasoning community. Therefore, this work focused on problems from real-world applications. The Cyc KB has thousands of queries that have been created by knowledge engineers and programmers for various projects (e.g., Project HALO [Friedland *et al.* 2004], HPKB project [Cohen *et al.* 1998]) and for testing the question-answering capability of the system. These queries are of varying levels of difficulty: some of them need just one transformation, others required the inference engine to back-chain on heavily used predicates that can lead to huge fan-out and high search

---

[7] Missing feature values are ignored during normalization, and then set to zero during training. This ensures that they are minimally informative because they are equal to the mean of the distribution [Hutter *et al.* 2014].

[8] We have experimented with forward, backward and exhaustive subset selection methods. All three methods lead to very similar set of selected features.

cost. We ensured that queries of both types were well represented in our test sets[9]. Based on the terms mentioned in them, the queries were divided into three test sets: (i) Test Set 1: Military and asymmetrical warfare domain, (ii) Test Set 2: Biology domain, and (iii) Test Set 3: Others (e.g., commonsense queries). The English translation of a query from test set 2 is shown below:

What causes the decline in MPF activity in M Phase?

The question shown above would lead to the query[10] (causes-SitTypeSitType ?cause MPFActivityDroppingInMPhase).

| Test Set | Method | No. of Queries | % Answ-ered | Q/A Imp. (%) | Time (hours) | Speedup |
|---|---|---|---|---|---|---|
| 1 | Baseline | 307 | 48 | - | 13.7 | - |
|   | DT | 307 | 57 | 19 | 11.7 | 1.17 |
|   | SL | 307 | 85 | 77 | 4.3 | 3.18 |
|   | DT+SL | 307 | 99 | 100 | 0.8 | 16.61 |
| 2 | Baseline | 1705 | 64 | - | 55.0 | - |
|   | DT | 1705 | 94 | 46 | 7.6 | 7.20 |
|   | SL | 1705 | 94 | 46 | 8.8 | 6.24 |
|   | DT+SL | 1705 | 96 | 50 | 4.0 | 13.50 |
| 3 | Baseline | 1736 | 26 | - | 100.8 | - |
|   | DT | 1736 | 81 | 211 | 22.6 | 4.45 |
|   | SL | 1736 | 86 | 230 | 17.5 | 5.76 |
|   | DT+SL | 1736 | 92 | 253 | 6.8 | 14.69 |

**Table 3: Experimental Results**

Recall that the inference engine uses a set of heuristics for ordering nodes during search, and the net score of a node can be written as $f(s) = w_0 + w_1.f_{DT}(s) + w_2.f_{SL}(s)$. Here, $f_{DT}(s)$ and $f_{SL}(s)$ refer to the scores returned by decision tree and statistical learning models discussed above. The first term, $w_0$, is the score returned by heuristics not discussed in this paper. In Table 3, the "baseline" version is obtained by setting both $w_1$ and $w_2$ to zero. By setting $w_2$ to 0, we can assess the efficacy of decision tree heuristics (rows labeled "DT" in Table 3). Similarly, we can study the utility of statistical learning models by setting $w_1$ to 0 (rows labeled "SL" in Table 3). The net contribution of both methods is shown in rows labeled "DT+SL." The experimental data was collected on a 4-core 3.40 GHz Intel processor with 32 GB of RAM. We used 18,383 decision trees, and ten best features identified by subset selection in these experiments. Due to the large time requirements of these queries, we restricted the cutoff time of each query to 5 minutes. Table 3 contains the results for three test sets. We see that both decision trees and multiple regression based models have led to significant speedups. The average speedup is a factor of 14. Since these heuristics steer the inference engine towards more productive parts of the search space, they improve question-answering (Q/A) performance too. The fifth column in Table 3 (labeled "Q/A Imp. (%)") shows the improvement in Q/A performance with respect to the baseline.

# 7 Conclusion

Deep deductive reasoning over large commonsense knowledge bases is critical for modern AI systems. The intractability of first-order logic has presented interesting research opportunities for understanding the causes of problem hardness and developing new algorithms for surmounting them. In this article, we have described two techniques to make reasoning more efficient. The first uses decision trees to guide the search toward germane rules by representing the *semantic* context in which a rule is expected to produce results. The second uses statistical regression techniques to provide an estimate of the number of answers a node is expected to provide based on *search meta-features*. The inference engine uses these heuristics to order nodes during search. Experimental results over thousands of queries show an order of magnitude speedup. These results suggest several lines of future work. First, we need to test these heuristics over even larger set of queries to understand their dynamics. Second, we want to extend our decision tree implementation to make probabilistic assessments. Next, we would like to experiment with other statistical models (e.g., regression splines, random forests) to improve the model quality. Finally, we believe that coupling this approach with a decision-theoretic model [Smith 1989, Greiner 1991] could yield a more complete theoretical model for making reasoning more efficient.

---

[9] The difficulty level of these queries can be gauged by looking at the average time requirements of the "baseline" version in Table 3. Initially some of the queries in our test set could not be answered in 20 minutes.

[10] (causes-SitTypeSitType *c e*) means that each instance of *c* is normally a cause of an instance of a situation type *e*.